\documentclass[10pt,twocolumn,letterpaper]{article}

\usepackage{cvpr}
\usepackage{times}
\usepackage{epsfig}
\usepackage{graphicx}
\usepackage{amsmath}
\usepackage{amssymb}
\usepackage{subcaption}
\usepackage{xcolor}
\usepackage{algorithm}
\usepackage{algpseudocode}
\usepackage{float}
\usepackage{makecell}

\algnewcommand\algorithmicinput{\textbf{Input:}}
\algnewcommand\algorithmicoutput{\textbf{Output:}}
\algnewcommand\Input{\item[\algorithmicinput]}%
\algnewcommand\Output{\item[\algorithmicoutput]}%

\usepackage{tabulary,multirow}

\newcolumntype{x}[1]{>{\centering\arraybackslash}p{#1pt}}

\newlength\savewidth\newcommand\shline{\noalign{\global\savewidth\arrayrulewidth
  \global\arrayrulewidth 1pt}\hline\noalign{\global\arrayrulewidth\savewidth}}

\makeatletter\renewcommand\paragraph{\@startsection{paragraph}{4}{\z@}
  {.5em \@plus1ex \@minus.2ex}{-.5em}{\normalfont\normalsize\bfseries}}\makeatother

\usepackage{array}
\newcolumntype{L}[1]{>{\raggedright\let\newline\\\arraybackslash\hspace{0pt}}m{#1}}
\newcolumntype{C}[1]{>{\centering\let\newline\\\arraybackslash\hspace{0pt}}m{#1}}
\newcolumntype{R}[1]{>{\raggedleft\let\newline\\\arraybackslash\hspace{0pt}}m{#1}}

\usepackage[pagebackref=true,breaklinks=true,colorlinks,bookmarks=false]{hyperref}

\cvprfinalcopy 

\def\cvprPaperID{5796} 

\ifcvprfinal\fi
\begin{document}

\title{Context-Aware Visual Compatibility Prediction}

\author{Guillem Cucurull\\
Element AI\\
{\tt\small gcucurull@elementai.com}
\and
Perouz Taslakian\\
Element AI\\
{\tt\small perouz@elementai.com}
\and
David Vazquez\\
Element AI\\
{\tt\small dvazquez@elementai.com}
}

\maketitle

\begin{abstract}
How do we determine whether two or more clothing items are \emph{compatible} or visually appealing?
Part of the answer lies in understanding of visual aesthetics, and is biased by personal preferences shaped by social attitudes, time, and place. In this work we propose a method that predicts compatibility between two items based on their visual features, as well as 
their \emph{context}. We define context as the products that are known to be compatible with each of these item. Our model is in contrast to other metric learning approaches that rely on pairwise comparisons between item features alone. We address the compatibility prediction problem using a graph neural network that learns to generate product embeddings conditioned on their context. We present results for two prediction tasks (\emph{fill in the blank} and \emph{outfit compatibility}) tested on two fashion datasets \emph{Polyvore} and \emph{Fashion-Gen}, and on a subset of the \emph{Amazon} dataset; we achieve state of the art results when using context information and show how test performance improves as more context is used.

\end{abstract}


\section{Introduction}
\label{sec:intro}

Predicting \emph{fashion compatibility} refers to the task of determining whether a set of fashion items go well together. In its ideal form, it involves understanding the visual styles of garments, being cognizant of social and cultural attitudes, and making sure that when worn together the outfit is aesthetically pleasing. The task is fundamental to a variety of industry applications such as personalized fashion design~\cite{kang2017visually}, outfit composition~\cite{feng2018interpretable}, wardrobe creation~\cite{hsiao2018creating}, item recommendation~\cite{shih2018compatibility} and fashion trend forecasting~\cite{al2017fashion}. Fashion compatibility, however, is a complex task that depends on subjective notions of style, context, and trend -- all properties that may vary from one individual to another and evolve over time.

\begin{figure}[t]
\begin{center}
\includegraphics[width=0.9\linewidth]{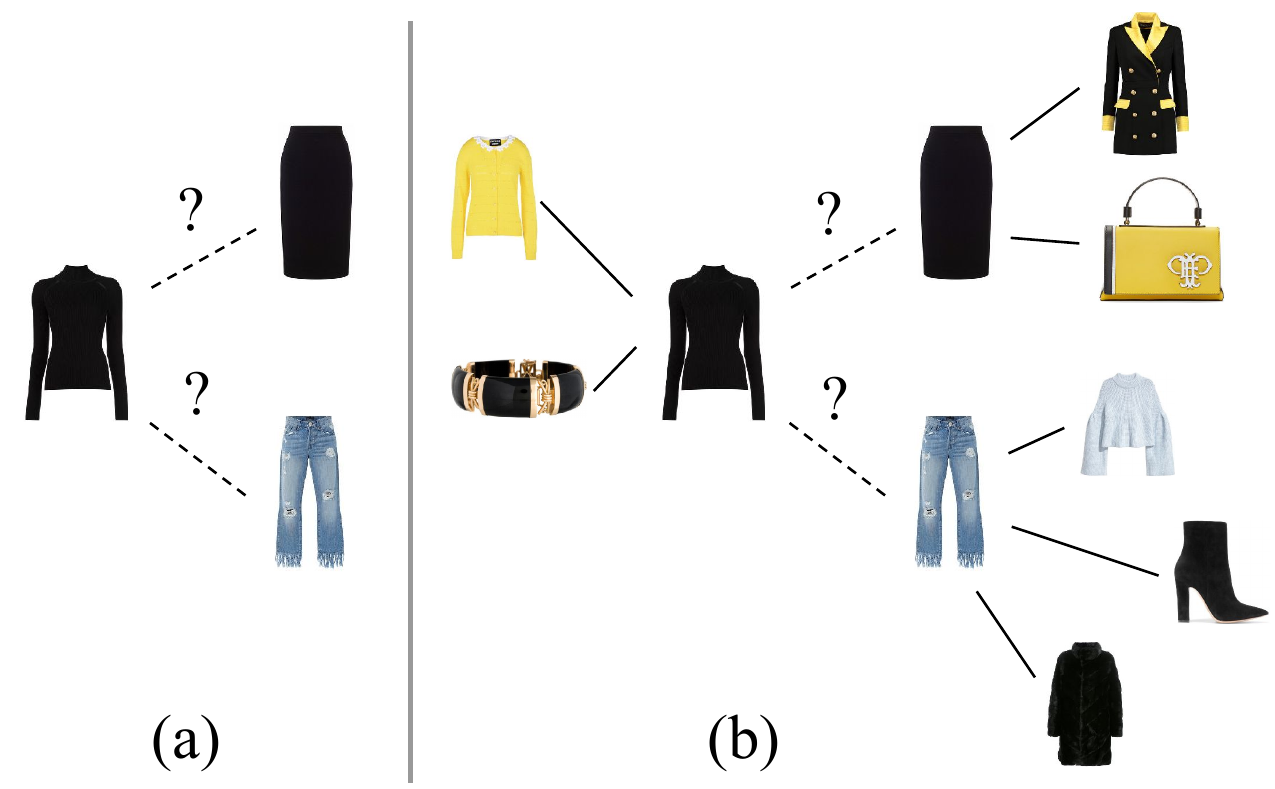}
\end{center}
\caption{\textbf{Fashion compatibility.} We use context information around fashion items to improve the task of fashion compatibility prediction. (a) standard methods compare pairs of items (b) we use a graph to exploit relational information to know the context of the items.}
\label{fig:main}
\end{figure}

Previous work~\cite{mcauley2015image, vasileva2018learning} on the problem of fashion compatibility prediction uses models that mainly perform pairwise comparisons between items based on item information such as image, category, description, \dots, etc. 
These approaches have the drawback that each pair of items considered are treated independently, making the final prediction rely on comparisons between the features of each item in isolation. 
In such a comparison mechanism that discards context, the model makes the same prediction for a given pair of clothing items every time.
For example, if the model is trained to match a specific style of shirt with a specific style of shoes, it will consistently make this same prediction every time. However, as compatibility is a subjective measure that can change with trends and across individuals, such inflexible behaviour is not always desirable at test time.
The compatibility between the aforementioned shirt and shoes is not only defined by the features of these items alone, but is also biased by the individual's preferences and sense of fashion. We thus define the \emph{context} of a clothing item to be the set of items that it is compatible with, and address the limitation of inflexible predictions by introducing a model that makes compatibility decisions based on the visual features, as well as the context of each item.
This consideration gives the model some background as to what we consider ``compatible'', in itself a subjective bias of the individual and the trend of the time.


In this paper, we propose to leverage the underlying relational information between items in a collection to make better compatibility predictions. We use fashion as our theme, and 
represent clothing items and their pairwise compatibility as a graph, where vertices are the fashion items and edges connect pairs of items that are compatible; we then use a graph neural network based model to learn to predict edges. Our model is based on the graph auto-encoder framework~\cite{kipf2016variational}, which defines an encoder that computes node embeddings and a decoder that is applied on the embedding of each product. Graph auto-encoders have previously been used for related problems such as recommender systems~\cite{vdberg2017graph}, and we extend the idea to the fashion compatibility prediction task. The encoder part of the model computes item embeddings depending on their connections, while the decoder uses these embeddings to compute the compatibility between item pairs. By conditioning the embeddings of the products on the neighbours, the style information contained in the representation is more robust, and hence produces more accurate compatibility predictions. 
This accuracy is tested by a set of experiments we perform on three datasets: Polyvore~\cite{han2017learning}, Fashion-Gen~\cite{rostamzadeh2018fashion} and Amazon~\cite{mcauley2015image}, and through two tasks (1) outfit completion (see Section~\ref{sssec:fitb}) and (2) outfit compatibility prediction (see Section~\ref{sssec:compat}). 
We compare our model with previous methods and obtain state of the art results. 
During test time, we provide our model with varying amount of context of each item being tested
and empirically show, in addition, that the more context we use, the more accurate our predictions get. 



This work has the following main contributions, (1) we propose the first fashion compatibility method that uses context information; (2) we perform an empirical study of how the amount of neighbourhood information used during test time influences the prediction accuracy; and (3) we show that our method outperforms other baseline approaches that do not use the context around each item on the Polvvore~\cite{han2017learning}, Fashion-Gen~\cite{rostamzadeh2018fashion}, and Amazon~\cite{mcauley2015image} datasets.

\begin{figure*}[t]
\begin{center}
\includegraphics[width=1.0\textwidth]{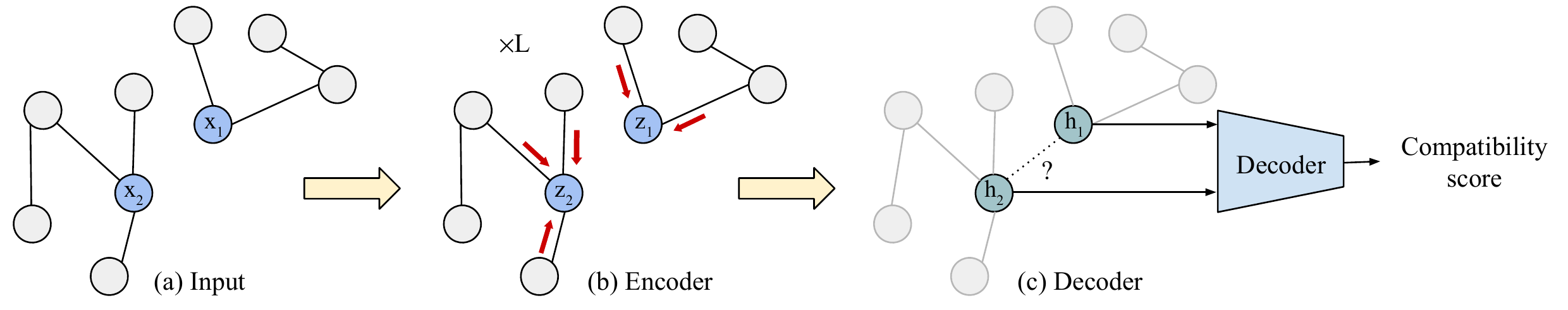}
\end{center}
\caption{\textbf{Method.} We pose fashion compatibility as an edge prediction problem. Our method consists of an encoder, which computes new embeddings for each product depending on their connections, and a decoder that predicts the compatibility score of two items. (a) Given the nodes $x_1$ and $x_2$ we want to compute their compatibility. (b) The encoder computes the embeddings of the nodes by using $L$ graph convolutional layers that merge information from their neighbours. (c) The decoder computes the compatibility score using the embeddings computed with the encoder.}
\label{fig:model}
\end{figure*}

\section{Related Work}
\label{sec:related_work}

As our proposed model uses graph neural networks to perform fashion compatibility prediction,
we group previous work related to our proposed model into two categories that we discuss in this section. 
In what follows, an \emph{outfit} is a set of clothing items that can be worn concurrently.
We say that an outfit is \emph{compatible}, if the clothing items composing the outfit are aesthetically pleasing when worn together;
we \emph{extend} an outfit when we add clothing item(s) to the set composing the outfit. 

\paragraph{Visual Fashion Compatibility Prediction.} 

To approach the task of visual compatibility prediction, McAuley \emph{et al.}~\cite{mcauley2015image} learn a compatibility metric on top of CNN-extracted visual features, and apply their method to pairs of products such that the learned distance in the embedding space is interpreted as compatibility. Their approach is improved by Veit \emph{et al.}~\cite{veit2015learning}, who instead of using pre-computed features for the images, use an end-to-end siamese network to predict compatibility between pairs of images. A similar end-to-end approach~\cite{kang2017visually} shows that jointly learning the feature extractor and the recommender system leads to better results. The evolution of fashion style has an important role in compatibility estimation, and He \emph{et al.}~\cite{he2016ups} study how previous methods can be adapted to model the visual evolution of fashion trends within recommender systems.

Some variations of this task include predicting the compatibility of an outfit, to generate outfits from a personal closet~\cite{tangseng2017recommending} for example, or determining the item that best extends a partial outfit. To approach these tasks, Han \emph{et al.}~\cite{han2017learning} consider a fashion outfit to be an \emph{ordered} sequence of products and use a bidirectional LSTM on top of the CNN-extracted features from the images and semantic information extracted from text in the embedding space. This method was improved by adding a new style embedding for the full outfit~\cite{nakamura2018outfit}. Vasileba \emph{et al.}~\cite{vasileva2018learning} also use textual information to improve the product embeddings, along with using conditional similarity networks~\cite{veit2017conditional} to produce type-conditioned embeddings and learn a metric for compatibility. This approach projects each product embedding to a new space, depending on the type of the item pairs being compared. 

\paragraph{Graph Neural Networks.} Extending neural networks to work with graph structured data was first proposed by Gori \emph{et al.}~\cite{gori2005new} and Scarselli \emph{et al.}~\cite{scarselli2009graph}. The interest in this topic resurged recently, with the proposal of spectral graph neural networks~\cite{bruna2013spectral} and its improvements~\cite{defferrard2016convolutional, kipf2016semi}. 
Gilmer \emph{et al.}~\cite{gilmer2017neural} showed that most of the methods that apply neural networks to graphs~\cite{monti2017geometric, velickovic2017graph, hamilton2017inductive} can be seen as specific instances of a learnable message passing framework on graphs. For an in-depth review of different approaches that apply neural networks to graph-structured data, we refer the reader to the work by Bronstein \emph{et al.}~\cite{bronstein2017geometric} and Battaglia \emph{et al.}~\cite{battaglia2018relational}, which explores how relational inductive biases can be injected in deep learning architectures. 

Graph neural networks have been applied to product recommendation, which is similar to product compatibility prediction. In this task, the goal is to predict compatibility between users and products (as opposed to a pair of products). Van den Berg \emph{et al.}~\cite{vdberg2017graph} showed how this task can be approached as a link prediction problem in a graph. Similarly, graphs can also be used to take advantage of the structure within the rows and columns of a matrix completion problem applied to product recommendation~\cite{kalofolias2014matrix, monti2017geometric_mc}. Recently, a graph-based recommender system has been scaled to web-scale~\cite{ying2018graph}, operating on a graph with more than 3 billion nodes consisting of pins and boards from Pinterest.





\section{Proposed Method}
\label{sec:method}

The approach we use in this work is similar to the metric learning idea of Vasileba \emph{et al.}~\cite{vasileva2018learning}, but rather than using text to improve products embeddings, we use a graph to exploit structural information and obtain better product embeddings. 
Our model is based on the graph auto-encoder (GAE) framework defined by Kipf \emph{et al.}~\cite{kipf2016variational}, which has been used for tasks like knowledge base completion~\cite{schlichtkrull2018modeling} and collaborative filtering~\cite{vdberg2017graph}.
In this framework, the encoder gets as input an incomplete graph, and produces an embedding for each node. Then, the node embeddings are used by the decoder to predict the missing edges in the graph.

Let $\mathcal{G} = (\mathcal{V}, \mathcal{E})$ be an undirected graph with $N$ nodes $i \in \mathcal{V}$ and edges $(i, j) \in \mathcal{E}$ connecting pairs of nodes. Each node in the graph is represented with a vector of features $\vec{x_i} \in \mathbb{R}^F$, and $\boldsymbol{X} = \{\vec{x_0}, \vec{x_1}, \dots, \vec{x}_N-1 \}$ is a $\mathbb{R}^{N \times F}$ matrix that contains the features of all nodes in the graph. Each row of $\boldsymbol{X}$, denoted as $\boldsymbol{X}_{i,:}$, contains the features of one node, \textit{i.e.} $\boldsymbol{X}_{i, 0}, \boldsymbol{X}_{i, 1}, \dots, \boldsymbol{X}_{i, N-1}$ represent the features of the $i^{th}$ node. 
The graph is represented by an adjacency matrix $\boldsymbol{A} \in \mathbb{R}^{N \times N}$,
where $\boldsymbol{A}_{i,j} = 1$ if there exist an edge between nodes $i$ and $j$ and $\boldsymbol{A}_{i,j} = 0$ otherwise.

The objective of the model is to learn an encoding $\boldsymbol{H} = f_{enc}(\boldsymbol{X}, \boldsymbol{A})$ and a decoding ${\boldsymbol{A} = f_{dec}(\boldsymbol{H})}$ function. The \emph{encoder} transforms the initial features~$\boldsymbol{X}$ into a new representation $\boldsymbol{H} \in \mathbb{R}^{N \times F'}$, depending on the structure defined by the adjacency matrix $\boldsymbol{A}$. This new matrix follows the same structure as the initial matrix~$\boldsymbol{X}$, so the $i$-th row $\boldsymbol{H}_{i, :}$ contains the new features for the $i$-th node. Then, the \emph{decoder} uses the new representations to reconstruct the adjacency matrix. This whole process can be seen as \emph{encoding} the input features to a new space, where the distance between two points can be mapped to the probability of whether or not an edge exists between them.
We use a \emph{decoder} to compute this probability using the features of each node: $p((i, j) \in \mathcal{E}) = f_{dec}(\boldsymbol{H}_{i,:}, \boldsymbol{H}_{j,:})$, which for our purposes represents the \emph{compatibility} between items $i$ and $j$.


In this work, the \emph{encoder} is a Graph Convolutional Network (Section \ref{ssec:encoder}) and the \emph{decoder} (Section \ref{ssec:decoder}) learns a metric to predict the compatibility score between pairs of products $(i,j)$. Figure \ref{fig:model} shows a scheme of how this encoder-decoder mechanism works.

\subsection{Encoder}
\label{ssec:encoder}
From the point of view of a single node $i$, the encoder will transform its initial visual features $\vec{x}_i$ into a new representation $\vec{h}_i$. The initial features, which can be computed with a CNN as a feature extractor, contain information about how an item looks like, \textit{e.g.}, shape, color, size. However, we want the new representation produced by the encoder to capture not only the product properties but also structural information about the other products it is compatile with. In other words, we want the new representation of each node to contain information about itself, but also about its neighbours $\mathcal{N}_i$, where $\mathcal{N}_i = \{j \in \mathcal{V}\ | \boldsymbol{A}_{i,j} = 1\}$ denotes the set of nodes that are connected to node $i$. Therefore, the encoder is a function that \emph{aggregates} the local neighbourhood around a node $\vec{h}_i = f_{enc}(\vec{x}_i, \mathcal{N}_i) : \mathbb{R}^F  \rightarrow \mathbb{R}^{F'}$ to include neighbourhood information in the learned representations. This function is implemented as a deep Graph Convolutional Network (GCN)~\cite{kipf2016semi} that can have several hidden layers. 
Thus, the final value of $\vec{h}_i$ is a composition of the functions computed at each hidden layer, which produces hidden activations $\vec{z}_i^{\,(l)}$. A single layer takes the following form.

\begin{equation}
    \vec{z}_i^{\,(l+1)} = ReLU \left( \vec{z}_i^{\,(l)} \boldsymbol{\Theta}_0^{(l)} + \sum_{j \in \mathcal{N}_i} \frac{1}{|\mathcal{N}_i|} \vec{z}_j^{\,(l)} \boldsymbol{\Theta}_1^{(l)} \right)
    \label{eq:fneigh}
\end{equation}

Here, $\vec{z}_i^{\,(l)}$ is the input of the $i$-th node at layer $l$, and $\vec{z}_i^{\,(l+1)}$ is its output.
In its matrix form, the function operates on all the nodes of the graph at the same time:

\begin{equation}
    \boldsymbol{Z}^{(l+1)} = ReLU \left( \sum_{s=0}^S \tilde{\boldsymbol{A}_s}\boldsymbol{Z}^{(l)}\boldsymbol{\Theta}_s^{(l)} \right)
    \label{eq:gconv}
\end{equation}

Here, $\boldsymbol{Z}^{(0)} = \boldsymbol{X}$ for the first layer. We denote $\tilde{\boldsymbol{A}_s}$ as the normalized $s$-{th} step adjacency matrix, where $\boldsymbol{A}_0 = I_N$ contains self-connections, and $\boldsymbol{A}_1 = \boldsymbol{A} + I_N$ contains first step neighbours with self-connections. We let $\boldsymbol{\tilde{A}} = \boldsymbol{D}^{-1}\boldsymbol{A}$, normalizing it row-wise using the diagonal degree matrix $\boldsymbol{D}_{ii} = \sum_j \boldsymbol{A}_{i,j}$. 
Context information is controlled by the parameter $S$ that represents the \emph{depth} of the neighbourhood that is being considered during training: the neighbourhood at depth $s$ of node $i$ is the set of all nodes that are at distance (number of edges traveled) at most $s$ from $i$. We let $S=1$ for all our experiments, meaning that we only use neighbours at depth one in each layer. $\boldsymbol{\Theta}_s^{(l)}$ is a $\mathbb{R}^{F \times F'}$ matrix, which contains the trainable parameters for layer $l$. 
We apply techniques such as batch normalization \cite{ioffe2015batch}, dropout~\cite{srivastava2014dropout} or weight regularization at each layer.

Finally, we introduce a regularization technique applied to the matrix $\boldsymbol{A}$, which consists of randomly removing all the incident edges of some nodes with a probability $p_{drop}$. The goal of this technique is two-fold: (1) it introduces some changes in the structure of the graph, making it more robust against changes in structure, and (2) it trains the model to perform well for nodes that do not have neighbours, making it more robust to scenarios with low relational information. 

\begin{algorithm}[t]
\caption{Compatibility prediction between nodes}
\label{alg:method}
\begin{algorithmic}[1]
\Input
\Statex $\boldsymbol{X}$ - Feature matrix of the nodes
\Statex $\boldsymbol{A}$ - Adjacency matrix of nodes relations
\Statex $(i, j)$ - Pairs of nodes for assessing compatibility
\Output The compatibility score $p$ between nodes $i$ and $j$

\Statex ~
\State $L = 3$ \Comment{Use 3 graph convolutional layers}
\State $S = 1$ \Comment{Consider neighbours 1 step away}

\State $\boldsymbol{H}$  = \Call{Encoder}{$\boldsymbol{X}$, $\boldsymbol{A}$}

\State $p$  = \Call{Decoder}{$\boldsymbol{H}$, $i$, $j$}

\Statex ~

\Function{Encoder}{$\boldsymbol{X}$, $\boldsymbol{A}$}
  \State $\boldsymbol{A}_0, \boldsymbol{A}_1 = I_L, I_L+\boldsymbol{A}$
  \State $\boldsymbol{\tilde{A}_1} = \boldsymbol{D}^{-1}\boldsymbol{A}_1$ \Comment{Normalize the adj. matrix}
  \State $\boldsymbol{Z}^{(0)} = \boldsymbol{X}$
    \For {each layer $l=0,...,L-1$}
        \State $\boldsymbol{Z}^{(l+1)} = ReLU \left( \sum\limits_{s=0}^S \boldsymbol{\tilde{A}}_s\boldsymbol{Z}^{(l)}\boldsymbol{\Theta}_s^{(l)} \right)$
    \EndFor
  \State \Return $\boldsymbol{Z}^{(L)}$
\EndFunction
\Statex ~
\Function{Decoder}{$\boldsymbol{H}$, $i$, $j$}
  \State \Return $ \sigma \left(\left|\boldsymbol{H}_{i,:} - \boldsymbol{H}_{j,:}\right| \vec{\omega}^{T} + b\right)$
\EndFunction

\end{algorithmic}
\end{algorithm}

\subsection{Decoder}
\label{ssec:decoder}
We want the decoder to be a function that computes the probability that two nodes are connected. This scenario is known as metric learning~\cite{bellet2013survey}, where the goal is to learn a notion of similarity or compatibility between data samples. It is relevant to note that similarity and compatibility are not exactly the same. Similarity measures how similar two nodes are, for example two shirts might be similar because they have the same shape and color, but they are not necessarily compatible. Compatibility is a property that measures how well two items go together.

In its general form, metric learning can be defined as learning a function $d(\cdot, \cdot) : \mathbb{R}^N \times \mathbb{R}^N \rightarrow \mathbb{R}_0^+$ that represents the distance 
between two $N$-dimensional vectors. Therefore, our decoder function takes inspiration from other metric learning approaches~\cite{koch2015siamese, hoffer2015deep, snell2017prototypical}. In our case, we want to train the decoder to model the compatibility between pairs of items, so we want the output of $d(\cdot, \cdot)$ to be bounded by the interval~$[0,1]$. 

The decoder function we use is similar to the one proposed by~\cite{garcia2017few}. Given the representations of two nodes $\vec{h}_i$ and $\vec{h}_j$ computed with the \emph{encoder} model described above, the \emph{decoder} outputs the probability $p$ that these two nodes are connected by an edge.

\begin{equation}
    p = \sigma\left(\left|\vec{h}_i - \vec{h}_j\right| \vec{\omega}^{T} + b\right)
\label{eq:decoder}
\end{equation}

Here $\left| \cdot \right|$ is absolute value, and $\vec{\omega} \in \mathbb{R}^{F'}$ and $b \in \mathbb{R}$ are learnable parameters. $\sigma(\cdot)$ is the sigmoid function that maps a scalar value to a valid 
probability $\in 
(0,1)$.

The form of the decoder described in Equation~\ref{eq:decoder} can be seen as a logistic regression decoder operating on the absolute difference between the two input vectors. The absolute value is used to ensure that the decoder is symmetric, \textit{i.e.}, the output of $d(\vec{h}_i, \vec{h}_j)$ and $d(\vec{h}_j, \vec{h}_i)$ is the same, making it invariant to the order of the nodes.


\begin{figure*}[t]
\begin{center}
\begin{subfigure}{.28\textwidth}
  \centering
  \includegraphics[width=\linewidth]{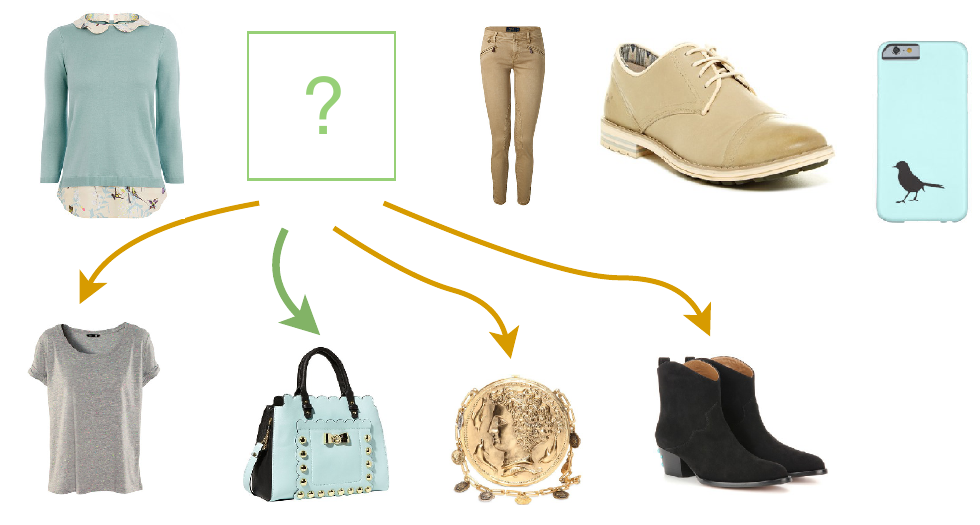}
  \caption{FITB example}
  \label{fig:fitb_examplea}
\end{subfigure}%
\hspace{1em}
\begin{subfigure}{.23\textwidth}
  \centering
  \includegraphics[width=\columnwidth]{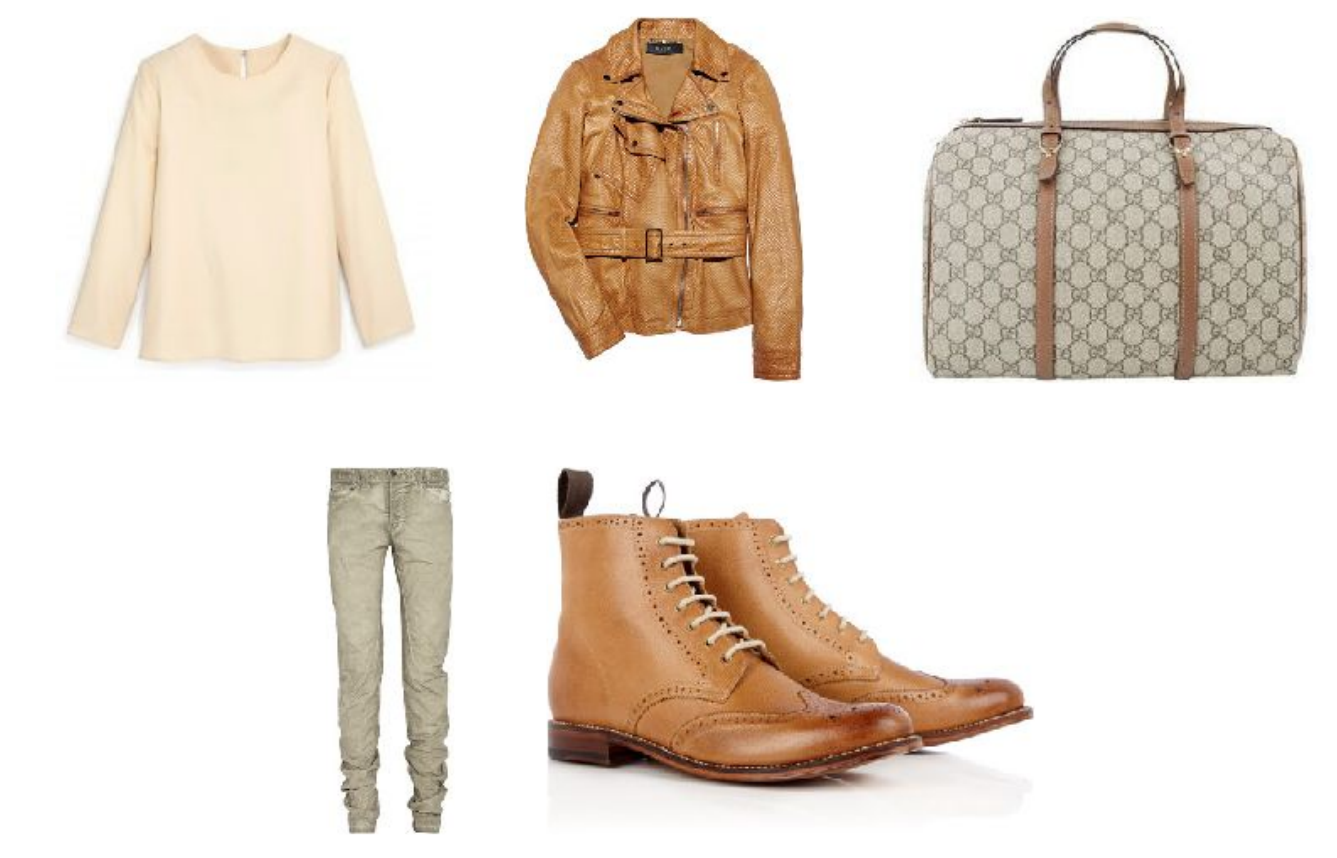}
  \caption{Valid outfit}
  \label{fig:valid_outfit}
\end{subfigure}%
\hspace{1em}
\begin{subfigure}{.23\textwidth}
  \centering
  \includegraphics[width=\linewidth]{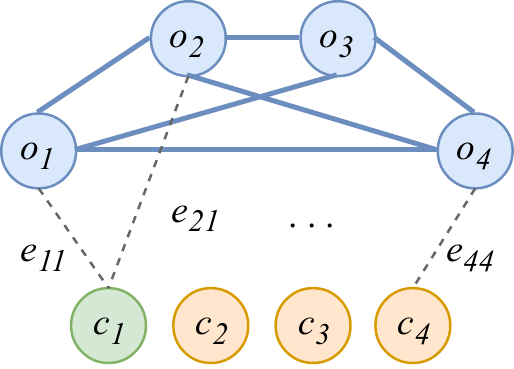}
  \caption{FITB as a graph}
  \label{fig:fitb_graph}
\end{subfigure}
\begin{subfigure}{.20\linewidth}
  \centering
  \includegraphics[width=\columnwidth]{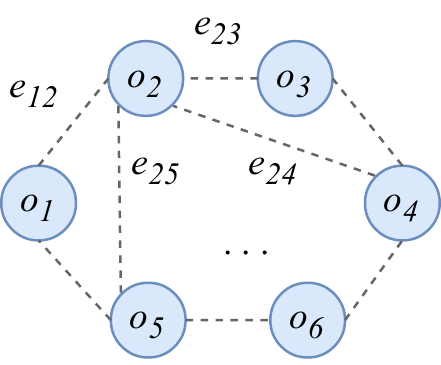}
  \caption{Compatibility as a graph}
  \label{fig:compat_graph}
\end{subfigure}
\end{center}
\caption{\textbf{Tasks.} We evaluate our model in two different tasks. (a) shows an example of a FITB question for the first task, and (b) shows an example of a valid outfit for the seconf task. (c) Shows how a FITB question can be posed as an edge prediction problem in a graph and (d) shows how the compatibility prediction for an outfit can be posed as an edge prediction problem.}
\end{figure*}

\subsection{Training}
\label{sseq:training}
The model is trained to predict compatibility among the products. With $\boldsymbol{A}$ being the adjacency matrix of the graph of items, we randomly remove a subset of edges to generate an incomplete adjacency matrix~$\boldsymbol{\hat{A}}$. The set of edges removed is denoted by~$\mathcal{E}^+$, as they represent \textit{positive} edges, \textit{i.e.}, pairs of nodes $(i,j)$ such that $\boldsymbol{A}_{i,j} = 1$. We then randomly sample a set of \textit{negative} edges $\mathcal{E}^-$, which represent pairs of nodes $(i, j)$ that are not connected, \textit{i.e.}, products that are not compatible. The model is trained to predict the edges $\mathcal{E}_{train} = (\mathcal{E}^+, \mathcal{E}^-)$ that contain both positive and negative edges. Therefore, given the incomplete adjacency matrix $\boldsymbol{\hat{A}}$ and the initial features for each node $\boldsymbol{X}$, the decoder predicts the edges defined in $\mathcal{E}_{train}$, and the model is optimized by minimizing the cross entropy loss between the predicted edges and their ground truth values, which is $1$ for the edges in $\mathcal{E}^+$ and $0$ for the edges in $\mathcal{E}^-$.

A schematic overview of the model can be seen in Figure~\ref{fig:model}, and Algorithm~\ref{alg:method} shows how to compute the compatibility between two products using the \emph{encoder} and \emph{decoder} described above.


\section{Experimental setup}
\label{sec:exp_setup}

\subsection{Tasks}
\label{ssec:tasks}
We apply our model to two tasks that can be recast as a graph edge prediction problem. In what follows, we let $\{o_1, \dotsc, o_{N-1}\}$ denote the set of $N$ fashion items in a given outfit, and $e_{i,j}$ denote the edge between nodes $i$ and $j$.

\paragraph{Fill In The Blank (FITB).}
\label{sssec:fitb}
The fill-in-the-blank task consists of choosing the item that best extends an outfit from among a given set of possible item choices. We follow the setup described in Han~\emph{et al.}~\cite{han2017learning}, where one FITB question is defined for each test outfit. Each question consists of a set of products that form a partial outfit, and a set of possible choices  $\{c_0, \dotsc, c_{M-1}\}$ that includes the correct answer and $M-1$ randomly chosen products. In our experiments we set the number of choices to 4. An example of one of these questions can be seen in Figure~\ref{fig:fitb_examplea}, where the top row shows the products of a partial outfit and the bottom row shows the possible choices for extending it. FITB can be framed as an edge prediction problem where the model first generates the probability of edges between item pairs $(o_i, c_j)$ for all $i=0,\dotsc, N-1$ and $j=0,\dotsc,M-1$. Then, the score for each of the $j$ choices is computed as $\sum_{i=0}^{N-1} e_{i,j}$, and the one with the highest score is the item that is selected to be added to the partial outfit. The task itself is evaluated using the same metric defined by Han~\emph{et al.}~\cite{han2017learning}: by measuring whether or not the correct item was selected from the list of choices.
%

\paragraph{Outfit Compatibility Prediction.}
\label{sssec:compat}
In the outfit compatibility prediction task, the goal is to produce an outfit \emph{compatibility score}, which represents the overall compatibility of the items forming the outfit. Scores close to $1$ represent compatible outfits, and scores close to $0$ represent incompatible outfits. The task can be framed as an edge prediction problem where the model predicts the probability of every edge between all possible item pairs; this means predicting the probability of ${\frac{N(N-1)}{2}}$ edges for each outfit. The compatibility score of the outfit is the average over all pairwise edge probabilities $\frac{2}{N(N-1)} \sum\limits_{i=0}^{N-1} \sum\limits_{j=i+1}^{N-1} e_{i,j}$.
The outfit compatibility prediction task is evaluated using the area under the ROC curve for the predicted scores.
%
%


\subsection{Evaluation by neighbourhood size}
\label{ssec:evaluation}
\begin{figure}[t]
\begin{center}
\begin{subfigure}{.33\columnwidth}
  \centering
  \includegraphics[width=\linewidth]{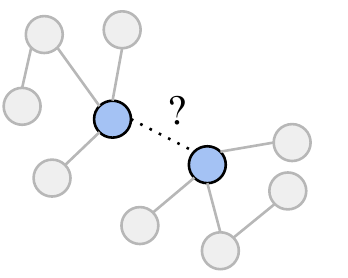}
  \caption{k=0}
  \label{fig:k0}
\end{subfigure}%
\begin{subfigure}{.33\columnwidth}
  \centering
  \includegraphics[width=\linewidth]{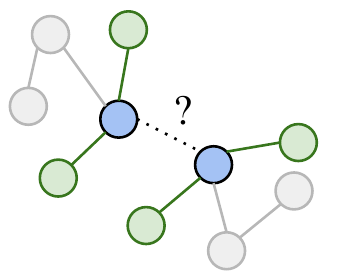}
  \caption{k=2}
  \label{fig:k2}
\end{subfigure}%
\begin{subfigure}{.33\columnwidth}
  \centering
  \includegraphics[width=\linewidth]{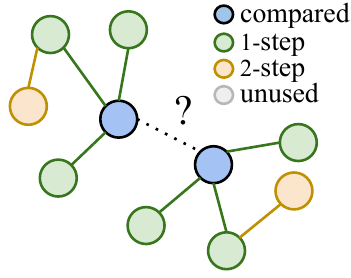}
  \caption{k=4}
  \label{fig:k4}
\end{subfigure}%
\end{center}
\caption{\textbf{Evaluation by $k$-neighbourhood.} BFS expansion of $k$ neighbours around two nodes. When (a) $k=0$ no neighbourhood information is used; (c) $k=4$ up to $4$ neighbourhood nodes are used for compatibility prediction.}
\end{figure}

%
%
Let the \emph{$k$-neighbourhood} of node $i$ in our relational graph be the set of $k$ nodes that are visited by a breadth-first-search process, starting from $i$. In order to measure the effect of the size of relational structure around each item, during testing we let each test sample contain the items and their $k$-neighbourhoods, and we evaluate our model by varying $k$.
Thus, when $k=0$ (Figure~\ref{fig:k0}) no relational information is used, and the embedding of each product is based only on its own features. As the value of $k$ increases (Figures~\ref{fig:k2} and \ref{fig:k4}), the embedding of the items compared will be conditioned on more neighbours. 
Note that this is applied only at evaluation time; during training, we use all available edges. For all results in the following sections we report the value of $k$ used for each experiment.

\subsection{Datasets}
\label{ssec:datasets}
We test our model on three datasets, as well as on a few of their variations that we discuss below.

\paragraph{The Polyvore dataset.}
\label{sssec:poly}
The Polyvore dataset~\cite{han2017learning} is a crowd-sourced dataset created by the users of a website of the same name; the website allowed its members to upload photos of fashion items, and collect them into outfits. 
It contains a total of $164{,}379$ items that form $21{,}899$ different outfits. 
The maximum number of items per outfit is $8$, and the average number of items per outfit is $6.5$. The graph is created by connecting each pair of nodes that appear in the same outfit with an edge. 
We train our model with the train set of the Polyvore dataset, and test it on a few variations obtained from this dataset, described below. 




The FITB task contains $3{,}076$ questions and the outfit compatibility task has $3{,}076$ valid, and $4{,}000$ invalid outfits. In the \emph{original} Polyvore dataset, the wrong FITB choices and the invalid outfits are selected randomly from among all remaining products. 
The \emph{resampled} dataset proposed by Vasileba~\emph{et al.}~\cite{vasileva2018learning} is more challenging: the incorrect choices in each question of the FITB task are sampled from the items having the same category as the correct choice; for outfit compatibility, outfits are sampled randomly such that each item in a given outfit is from a distinct category. 
We also propose a more challenging set which we call \emph{subset} where we limit the outfits size to 3 randomly selected items. In this scenario the tasks become harder because less information is available to the model.

\paragraph{The Fashion-Gen Outfits dataset.}
\label{sssec:fgen}
Fashion-Gen~\cite{rostamzadeh2018fashion} is a dataset of fashion products collected from an online platform that sells luxury goods from independent designers. Each product has images, descriptions, attributes, and relational information. 
Fashion-Gen relations are defined by professional designers and adhere to a general theme, while Polyvore's relations are generated by users with different tastes and notions of compatibility.

We created outfits from Fahion-Gen by grouping between $3$ and $5$ products that are connected together. The training set consists of $60{,}159$ different outfits from the collections $2015-2017$, and the validation and test sets have $2{,}683$ and $3{,}104$ outfits respectively, from the $2014$ collection. The incorrect FITB choices and the invalid outfits for the compatibility task are randomly sampled items that satisfy gender and category restrictions, as in the case of the resampled Polyvore dataset.

\paragraph{Amazon products dataset.}
\label{sssec:amazon_dataset}
The Amazon products dataset~\cite{mcauley2015image, he2016ups} contains over $180$ million relationships between almost $6$ million products of different categories. In this work we focus on the clothing products, and we apply our method to the \emph{Men} and \emph{Women} categories. There are $4$ types of relationships between items: (1) users who viewed $A$ also viewed $B$; (2) users who viewed $A$ bought $B$; (3) users who bought $A$ also bought $B$; and (4) users bought $A$ and $B$ simultaneously. For the latter two cases, we make the assumption that the pair or items $A$ and $B$ are compatible and evaluate our model based on this assumption. 
We evaluate our model by predicting the latter two, since they indicate products that might be complementary~\cite{mcauley2015image}. We use the features they provide, which are computed with a CNN.

\subsection{Training details}
\label{ssec:training_details}
Our model has $3$ graph convolutional layers with $S=1$, $350$ units, dropout of $0.5$ applied at the input and batch normalization at its output. The value of $p_{drop}$ applied to $\boldsymbol{A}$ is $0.15$. 
The input to each node are $2048$-dimensional feature vectors extracted with a ResNet-50~\cite{he2016deep} from the image of each product, and are normalized to zero-mean and unit variance. It is trained with Adam~\cite{kingma2014adam}, with a \textit{learning rate} of $0.001$ for $4,000$ iterations with early stopping.

The \emph{Siamese Network} baseline is trained with triplets of compatible and incompatible pairs of items. It consists on a ImageNet pretrained ResNet-50 at each branch and a metric learning output layer. We train it using SGD with a learning rate of $0.001$ and a momentum of $0.9$.

\section{Results}
\label{sec:results}

\begin{figure*}[t]
\begin{center}
\begin{subfigure}{.24\textwidth}
  \centering
  \includegraphics[width=\linewidth]{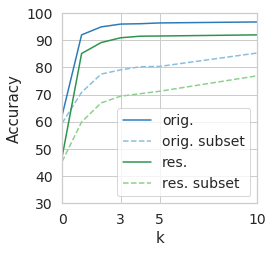}
  \caption{Polyvore FITB}
  \label{fig:k_fitb_poly}
\end{subfigure}%
\begin{subfigure}{.245\textwidth}
  \centering
  \includegraphics[width=\linewidth]{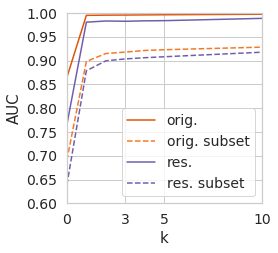}
  \caption{Polyvore compatibility}
  \label{fig:k_compat_poly}
\end{subfigure}%
\begin{subfigure}{.24\textwidth}
  \centering
  \includegraphics[width=\linewidth]{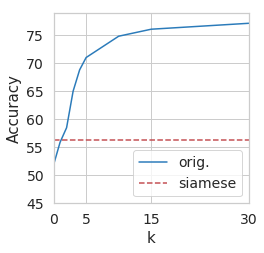}
  \caption{Fashion-Gen FITB}
  \label{fig:k_fitb_ssense}
\end{subfigure}%
\begin{subfigure}{.245\textwidth}
  \centering
  \includegraphics[width=\linewidth]{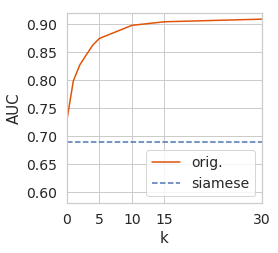}
  \caption{Fashion-Gen compatibility}
  \label{fig:k_compat_ssense}
\end{subfigure}
\end{center}
\caption{\textbf{Results.} Evaluation of our models for different values of $k$.}
\label{fig:k}
\end{figure*}

\subsection{Fill In The Blank}
\label{ssec:fitb}
\paragraph{Polyvore Original.} 
We report our results for this task in Table~\ref{tab:results-poly}. The first three rows correspond to previous work, and the following three rows show the scores obtained by our model for different values of $k$. As shown in the table, the scores consistently increases with $k$, from $62.2\%$ of accuracy with $k=0$ to $96.9\%$ with $k=15$. This behaviour is better seen in Figure~\ref{fig:k_fitb_poly} which shows how the accuracy in the FITB task increases as a function of $k$. When $k=0$ the other methods perform better, because without structure our model is simpler. However, we can see how as more neighbourhood information is used, the results in the FITB task increase, which shows that using information from neighbouring nodes is a useful approach if extra relational information is available.

\begin{table}[t]
\caption{\textbf{Polyvore Results.} Polyvore results for both the FITB and the compatibility prediction tasks. Resampled task is more difficult than the original one. \textsuperscript{\textdagger} Using only a subset of length $3$ of the original outfit.}
\vskip-0cm
\label{tab:results-poly}
\begin{center}
\begin{tabular}{l|cccc}
 & \multicolumn{2}{c}{\textbf{FITB Accuracy}} & \multicolumn{2}{c}{\textbf{Compat. AUC}}\\
\textbf{Method} & \textbf{Orig.} & \textbf{Res.} & \textbf{Orig.} &  \textbf{Res.}\\\shline
Siamese Net~\cite{vasileva2018learning}         & 54.2 & 54.4 & 0.85 & 0.85 \\
Bi-LSTM~\cite{han2017learning}                  & 68.6 & 64.9 & 0.90 & 0.94 \\
TA-CSN~\cite{vasileva2018learning}              & 86.1 & 65.0 & 0.98 & 0.93 \\\hline
Ours (k = 0)                                    & 62.2 & 47.0 & 0.86 & 0.76 \\
Ours (k = 3)                                    & 95.9 & 90.9 & 0.99 & 0.98 \\
Ours (k = 15)                                   & \textbf{96.9} & \textbf{92.7} & \textbf{0.99} & \textbf{0.99} \\\hline
Ours (k = 0)\textsuperscript{\textdagger}       & 59.5 & 45.3 & 0.69 & 0.64 \\
Ours (k = 3)\textsuperscript{\textdagger}       & 79.1 & 69.4 & 0.92 & 0.90 \\
Ours (k = 15)\textsuperscript{\textdagger}      & 88.2 & 82.1 & 0.93 & 0.92 \\
\end{tabular}
\end{center}
\end{table}

\paragraph{Polyvore Resampled.} 
For the resampled setup, the accuracy also increases with $k$, going from $47.0\%$ to $92.7\%$, which is lower than its original counterpart, showing that the resampled task is indeed more difficult.

\paragraph{Polyvore Subset.} 
The last rows of Table~\ref{tab:results-poly} (marked with \textdagger) correspond to this scenario, and we can see that compared to when using the full outfit, the the FITB accuracy drops from $96.9\%$ to $88.2\%$ for the original version, and from $92.7\%$ to $82.1\%$ for the resampled version, both at $k=15$.

\begin{table}[t]
\caption{\textbf{Fashion-Gen Results.} Results on the Fashion-Gen dataset for the FITB and compatibility tasks.}
\vskip-0cm
\label{tab:results-fgen}
\begin{center}
\begin{tabular}{l|cc}
\textbf{Method} & \textbf{FITB Acc.} & \textbf{Compatibility AUC}\\\shline
Siamese Network         & 56.3 &  0.69 \\     
 \hline
Ours (k = 0)            & 51.9 &  0.72  \\
Ours (k = 3)            & 65.0 &  0.84  \\
Ours (k = 15)           & 76.1 &  0.90  \\
Ours (k = 30)           & \textbf{77.1} &  \textbf{0.91}  \\
\end{tabular}
\end{center}
\end{table}

\paragraph{Fashion-Gen Outfits.}

The results for the FITB task on the Fashion-Gen dataset are shown in Table~\ref{tab:results-fgen} as a function of $k$. Similar to the results for variations of Polyvore, we see in Figure~\ref{fig:k_fitb_ssense} how an increase in the value of $k$ improves the performance of our model also for 
the Fashion-Gen dataset. For example, it increases by $20$ points by using up to $k=15$ neighbourhood nodes for each item, compared to using no neighbourhood information at all. When compared to the Siamese Network baseline, we observe how the siamese model is better than our model without structure, but with $k \geq 3$ our method outperforms the baseline.

\subsection{Outfit Compatibility Prediction} \label{sec:compat_results}

\paragraph{Polyvore Results.} 
Table~\ref{tab:results-poly} shows the results obtained by our model on the compatibility prediction task for different values of $k$. 
Similarly to the previous task, results show that using more neighborhood information improves the performance on the outfit compatibility task, where the 
AUC increases from $0.86$ with $k=0$ to $0.99$ with $k=15$.

\paragraph{Polyvore Resampled.} 
The scores on the resampled version are similar to the original version, increasing the AUC from $0.76$ to $0.99$ with a larger value for $k$.

\paragraph{Polyvore Subset.} The results on this test data is denoted with \textdagger ~in the table, and we see how in this scenario the scores decrease from $0.99$ to $0.93$ and $0.92$ for the original and resampled tasks respectively, both with $k=15$.
As with the FITB task, here we observe again how using extra information in the form of relations with other products is beneficial to achieve better performance. 

\paragraph{Fashion-Gen Outfits.} 
%
The results on this task for the Fashion-Gen outfits dataset are shown in the second column of Table~\ref{tab:results-fgen}, for different values of $k$. As can be seen, the larger the value of $k$, the better the performance. This trend is better shown in Figure~\ref{fig:k_compat_ssense}, where we can see how increasing $k$ from $0$ to $10$ steadily improves the performance, and plateaus afterwards.

\begin{figure}[t]
\begin{center}
\begin{subfigure}{.40\textwidth}
  \centering
  \includegraphics[width=\linewidth]{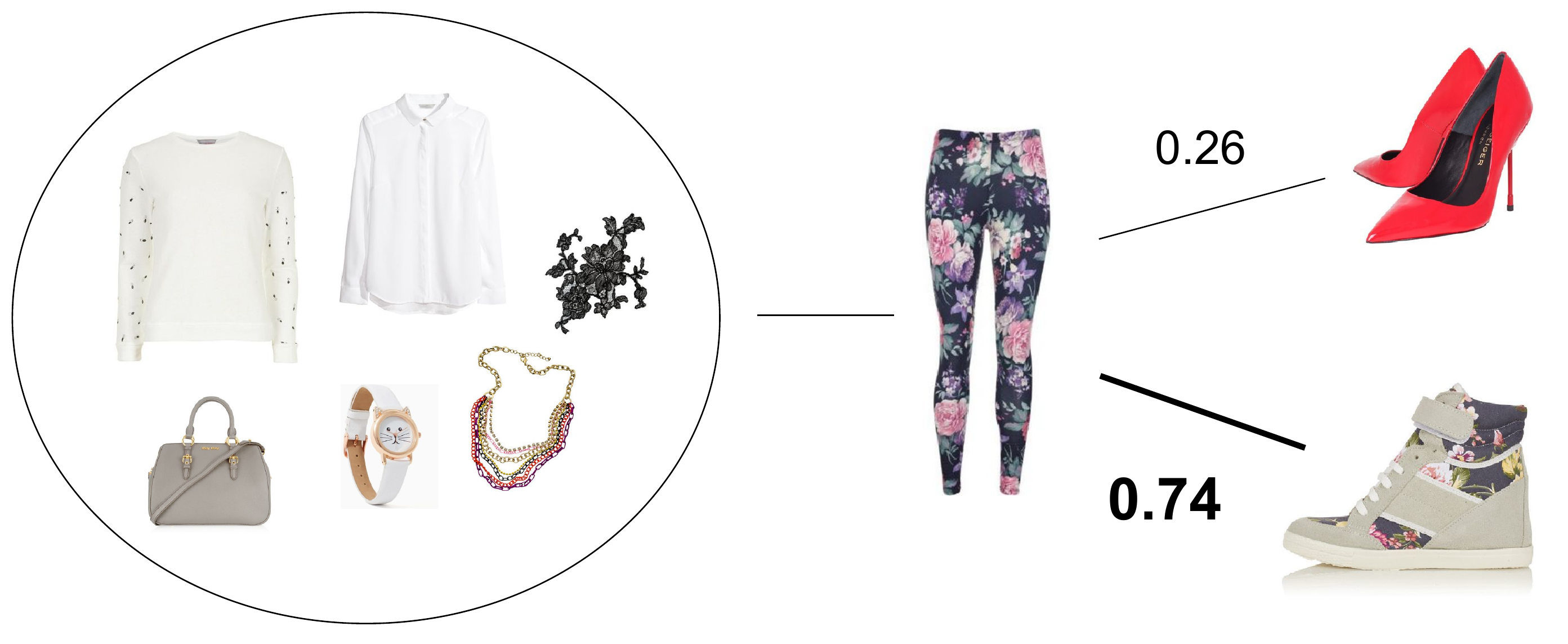}
  \caption{original context}
  \label{fig:context_a}
\end{subfigure}
\begin{subfigure}{.40\textwidth}
  \centering
  \includegraphics[width=\linewidth]{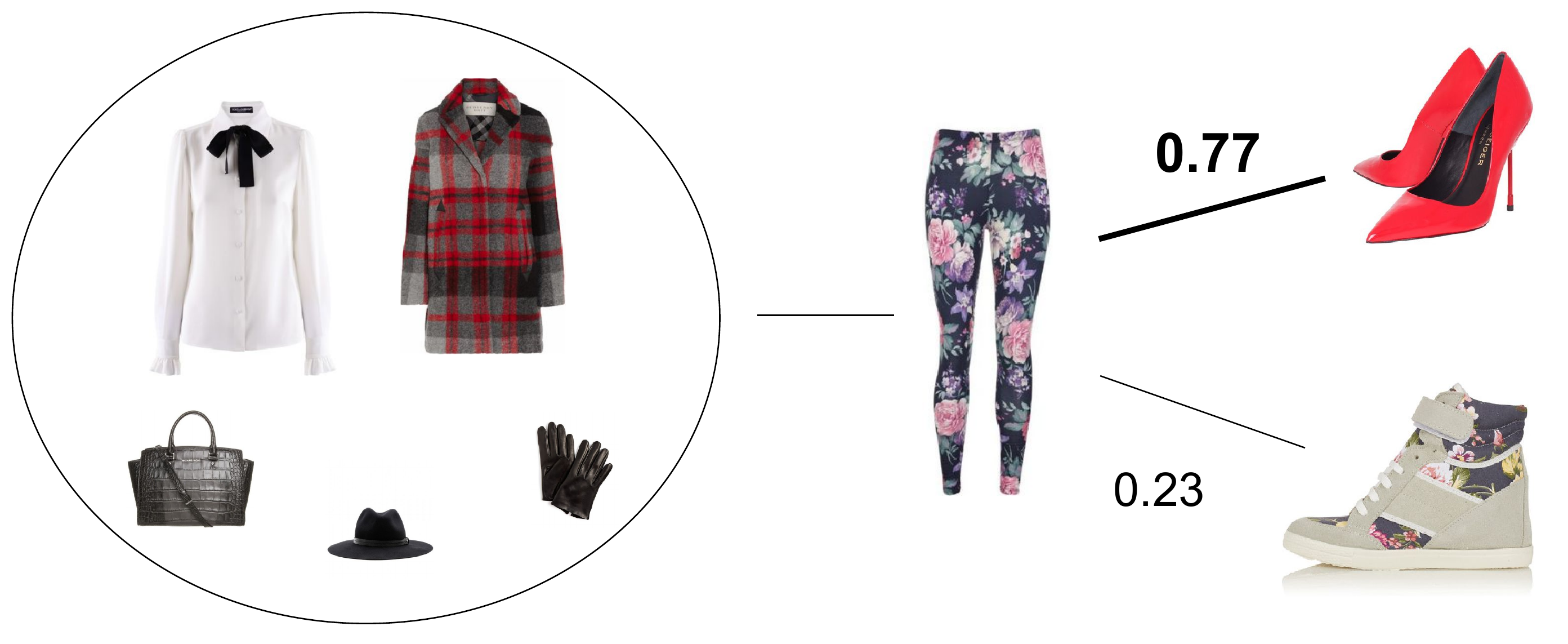}
  \caption{new context}
  \label{fig:context_b}
\end{subfigure}%
\end{center}
\caption{\textbf{Context matters.} (a) and (b) show how predicted compatibility between items depends on their context.}
\label{fig:context}
\end{figure}
\subsection{Context matters}
With the above experiments, we have seen how increasing the amount of neighbourhood information improves the results on all tasks. To better understand the role of context, 
we use an example from Polyvore to demonstrate how the context of an item can influence its predicted compatibility with another product. 
Figure \ref{fig:context} 
shows the compatibility predicted between a pair of trousers and two pairs of shoes depending on two different contexts. Figure~\ref{fig:context_a} shows the original context of the trousers, and the shoes selected are the correct ones. However, if we change the context of the trousers to a different set of clothes, as in Figure \ref{fig:context_b}, the outcome of the prediction is now a different pair of shoes (more formal one) that presumably 
are a better match given the new context.

\subsection{Amazon Links}
We also evaluate how our method can be applied to predict relations between products in the Amazon dataset. We train a model for each type of relationship and also evaluate how one model trained with clothes from one gender transfers to the other gender. This cross-gender setup allows us to evaluate how the model adapts to changes in context, as opposed to a baseline that ignores context altogether. In Table \ref{tab:results-amazon} we show that our model achieves state of the art results for the 'also bought' relation, and similar results for the 'bought together' relation. The 'bought together' relationship has much less connections than the 'also bought', so our model is less effective at using context to improve the results. However, since in that scenario the model has been trained with less connections, it performs better with $k=0$, because it is more similar to the training behaviour. In Table \ref{tab:results-gender-amazon} we show the results of one model trained with men's clothing and tested with women's clothing (and vice versa). The model denoted with \textdagger ~does not use relational information during training and testing, so is the baseline for not using contextual information ($k=0$). As it can be seen, the more neighbourhood information a model uses, the most robust it is to the domain change. This occurs because when the model relies on context, it can adapt better to unseen styles or clothing types.

\begin{table}[t]
\caption{\textbf{Amazon results.} Results on the Amazon dataset for the link prediction task.}
\vskip-0cm
\label{tab:results-amazon}
\begin{center}
\begin{tabular}{l|cccc}
 & \multicolumn{2}{c}{\textbf{Also bought}} & \multicolumn{2}{c}{\textbf{Bought together}}\\
\textbf{Method} & \textbf{Men} & \textbf{Women} & \textbf{Men} & \textbf{Women} \\\shline
McAuley \emph{et al.}~\cite{mcauley2015image} & 93.3 &  91.2 & \textbf{95.1} & 94.3\\
\hline
Ours (k = 0) & 57.9 &  53.8 & 79.5 & 71.7 \\
Ours (k = 3) & 92.6 &  92.9 & 94.5 & 94.5 \\
Ours (k = 10)  & \textbf{97.1} &  \textbf{95.8} & 94.0 & \textbf{94.8} \\
\end{tabular}
\end{center}
\end{table}

\begin{table}[t]
\caption{\textbf{Amazon cross-gender results.} Test the adaptability of the model by training and testing across different genders. Rows show the gender the model has been trained on, columns show the gender the model is tested with. \textsuperscript{\textdagger}~Model trained also with $k=0$ so it does not use context during training.}
\vskip-0cm
\label{tab:results-gender-amazon}
\begin{subtable}{.5\linewidth}
\centering
\scalebox{0.89}{
\tabcolsep=0.11cm
\begin{tabular}{lc|cc}
 & & \textbf{Men} & \textbf{Women} \\\shline
\multirow{2}{*}{k=0 \textsuperscript{\textdagger}} 
                    & Men & 95.0 & 58.3 \\
                    & Women & 66.5 & 93.2  \\
\hline
\multirow{2}{*}{k=0} & Men & 57.9 & 52.9 \\
                      & Women & 55.9 & 53.8  \\
\multirow{2}{*}{k=3} & Men & 92.6 & 79.8 \\
                      & Women & 86.5 & 92.9  \\
\multirow{2}{*}{k=10} & Men & 97.1 & 86.0 \\
                      & Women & 90.9 & 95.8  \\                      
\end{tabular}
}
\caption{\textbf{Also bought.}}
\end{subtable}%
\begin{subtable}{.5\linewidth}
\centering
\scalebox{0.89}{
\tabcolsep=0.11cm
\begin{tabular}{lc|cc}
 & & \textbf{Men} & \textbf{Women} \\\shline
\multirow{2}{*}{k=0 \textsuperscript{\textdagger}} 
                    & Men & 90.7 & 62.5 \\
                    & Women & 73.2 & 91.5  \\
\hline
\multirow{2}{*}{k=0} & Men & 79.5 & 61.8 \\
                      & Women & 68.5 & 71.7  \\
\multirow{2}{*}{k=3} & Men & 82.7 & 73.9 \\
                      & Women & 79.7 & 94.5  \\
\multirow{2}{*}{k=10} & Men & 94.0 & 74.3 \\
                      & Women & 83.2 & 94.8  \\                      
\end{tabular}
}
\caption{\textbf{Bought together.}}
\end{subtable}
\end{table}


\section{Conclusions}
In this paper we have seen how context information can be used to improve the performance on compatibility prediction tasks using a graph neural network based model. We experimentally show that increasing the amount of context improves the performance of our model on all tasks. We conduct experiments on three different fashion datasets and obtain state of the art results when context is used during test time. 


{\small
\bibliographystyle{ieee}
\bibliography{egbib}
}

\end{document}